\documentclass[conference]{IEEEtran}
\IEEEoverridecommandlockouts
\usepackage{amsmath,amssymb,amsfonts}
\usepackage{graphicx}
\usepackage{textcomp}
\usepackage{xcolor}
\usepackage{enumerate}
\usepackage{url}
\usepackage{mathtools}
\usepackage{comment}
\usepackage{bm}
\usepackage{bbm,color}
\usepackage{multirow}
\usepackage{tabularx}
\usepackage[T1]{fontenc}
\usepackage{aecompl}
\usepackage{algorithm}
\usepackage{algpseudocode}
\usepackage[numbers,sort&compress]{natbib}

\def\BibTeX{{\rm B\kern-.05em{\sc i\kern-.025em b}\kern-.08em
    T\kern-.1667em\lower.7ex\hbox{E}\kern-.125emX}}
\begin{document}

\title{Predictive Multi-level Patient Representations\\ from Electronic Health Records}
\author{\IEEEauthorblockN{Zichang Wang$^*$\thanks{$*$The two authors have equal contribution.}, Haoran Li$^*$, Luchen Liu, Haoxian Wu and Ming Zhang}
\IEEEauthorblockA{\textit{Department of Computer Science, Peking University, Beijing, China} \\
\{dywzc123, lhrshitc, liuluchen292\}@163.com, \{MOVIEGEORGE, mzhang\_cs\}@pku.edu.cn}
}

\maketitle

\begin{abstract}
The advent of the Internet era has led to an explosive growth in the Electronic Health Records (EHR) in the past decades. The EHR data can be regarded as a collection of clinical events, including laboratory results, medication records, physiological indicators, etc, which can be used for clinical outcome prediction tasks to support constructions of intelligent health systems. Learning patient representation from these clinical events for the clinical outcome prediction is an important but challenging step. Most related studies transform EHR data of a patient into a sequence of clinical events in temporal order and then use sequential models to learn patient representations for outcome prediction. However, clinical event sequence contains thousands of event types and temporal dependencies. We further make an observation that clinical events occurring in a short period are not constrained by any temporal order but events in a long term are influenced by temporal dependencies.  The multi-scale temporal property  makes it difficult for traditional sequential models to capture the short-term co-occurrence and the long-term temporal dependencies in clinical event sequences. In response to the above challenges, this paper proposes a Multi-level Representation Model (MRM). MRM first uses a sparse attention mechanism to model the short-term co-occurrence, then uses interval-based event pooling to remove redundant information and reduce sequence length and finally predicts clinical outcomes through Long Short-Term Memory (LSTM). Experiments on real-world datasets indicate that our proposed model largely improves the performance of clinical outcome prediction tasks using EHR data.
\end{abstract}


\section{Introduction}
In the past decades, the scale of Electronic Health Records (EHR) has exploded because of the advent of the Internet era, which makes the construction of electronic medical record systems possible. 

We focus on clinical event outcome prediction based on patient representation sequence learning.\cite{liu2019learning,liu2019early} The electronic medical record data can be considered as a collection of clinical events, including thousands of event types such as diagnosis, laboratory tests, medication records, activity records, and physical signs. The clinical outcome prediction based on patient representation sequence learns the low-dimensional representation of the patient from the electronic medical record data and predicts the results of the specified clinical events, which can assist the medical experts to make correct clinical decisions.

Some of the related works sort the clinical events in the electronic medical record data according to their time of occurrence and converted the electronic medical record data into a sequence of clinical events. On this basis, the embedded layer is used to represent the clinical events and then the sequence model is used to capture the temporal dependencies between events and predict the results of the specified clinical events. However, the clinical event outcome prediction model under this framework has several challenges: 
\begin{itemize}
    \item Clinical events in electronic medical records contain rich and complex high-dimensional information, which have thousands of types. 
    \item Events in a small neighborhood is out-of-order but the interaction of these events are also predictive. 
    \item The sequence is too long for sequence model like long short-term memory neural networks (LSTM) to capture the long-term dependency. 
\end{itemize}

To gently solve the above challenges, we propose a Multi-level Representation Model (MRM). MRM uses the attention mechanism to capture the short-term co-occurrence of the events and obtain a low-level neighborhood representation of events. The pooling mechanism is then used to reduce the length of the clinical event sequence based on the short-term out-of-order clinical events. Finally, MRM uses LSTM to capture long-term temporal dependencies between events, obtain final patient representation and predict the outcome of a given clinical event.

The main contributions of this paper are as follows:

\begin{itemize}
\item Compared to studies only using medical code or dozens of event types, this paper make
use of nearly a thousand event types and events' features to make predictions.
\item This paper proposes a multi-level representation model for patient medical records to capture the short-term co-occurrence and long-term temporal
dependencies between clinical events. The effect was verified in experiments with actual
data.
\item  The interval-based event pooling mechanism proposed in this paper preserves the integrity of
information while removing redundant information and reducing sequence length.
\end{itemize}

\section{Related Works}
\subsection{Patient Representation from EHR}
One general patient representation method to make direct use of high-dimensional EHR data is to use a vector that records the number of each type of clinical events to represent EHR data\cite{dai2016bagging, ghassemi14unfolding, huddar2016predicting}. However, it is obvious that it ignores the relative order of clinical events and lacks a more detailed description of the features of clinical events.

Another method is to use a matrix in which the rows of the matrix represent different time intervals and the columns of the matrix represent a type of event\cite{wang2013framework, zhou2014macro}. Wang et al. use the convolutional non-negative matrix factorization to resolve the matrices\cite{wang2013framework}. Zhou et al. decomposed the matrix into the product of Latent Medical Concept Matrix and Concept Value Evolution Matrix\cite{zhou2014macro}. These method depend on the time span which is set ahead and still only uses the event occurrence information and lacks more detailed features.

The Temporal Phenotyping proposed by Liu et al. converts the EHR data into a sequence diagram where the nodes represent clinical events and the edge weights represent the correlation of the connected nodes\cite{liu2015temporal}. Such method focuses on capturing the short-term co-occurrence of events and ignores the long-term temporal dependencies between important events.

\subsection{Deep Sequential models for EHR}
Some related works introduce the time information or the interval information of the events into the model to solve the problem of inconsistent sampling frequency\cite{bai2018interpretable, che2015deep, zheng2017capturing,liu2018learning}. For example, Che et al. multiply the hidden state by a time decay factor before calculating the next hidden state in Gated Recurrent Unit (GRU)\cite{che2015deep}. Zheng et al. balance the inheritance and update of hidden states based on the time decay function when updating the hidden layer state of GRU\cite{zheng2017capturing}. Bai et al. propose the Timeline model to model the decay rate of different events affecting patients\cite{bai2018interpretable}. These efforts use time decay factors to solve inconsistencies in clinical event sequences but do not consider the short-term out-of-order in clinical event sequence in EHR data. 

Choi et al. propose a model RETAIN\cite{choi2016retain} that combines RNN with attention mechanisms. RETAIN divides the sequence into several visits, and then uses attention mechanism to generate patient representation based on the visits. However, RETAIN uses only medical code information for clinical events. 

\section{Methodology}

\begin{figure}[!t]
\centering
\includegraphics[width=.47\textwidth]{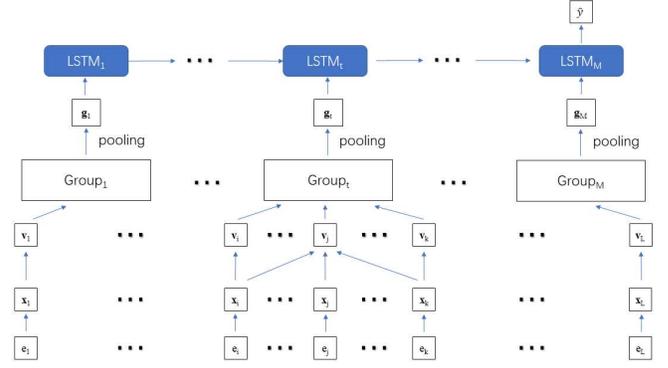}
\caption{The architecture of MRM: Event $e_i$ is encoded as a representation $\bm x_i$. The short-term co-occurrence mechanism gather the neighborhood information and generate the event representation $\bm v_i$. Then the interval-based event pooling mechanism divide the event sequence into groups and generate the group representation $\bm g_i$. Finally the event group representation sequence is fed to LSTM and generate the output $\hat{y}.$}
\label{fig:model}
\end{figure}

This chapter shows the Multi-level Representation Model (MRM) proposed in this paper in detail. This chapter formalizes the problems studied in this paper, and then introduces the mechanisms of the model showed in figure \ref{fig:model}.
\subsection{Notations}
A clinical event $e$ can be formalized as a quadruple $(code, t, catFea, numFea)$. We use $e.code$, $e.t$, $e.catFea$, $e.numFea$ standing for the encoding of the event, occurrence time, category feature and numerical feature.

\subsection{Multi-level Representation Model}
\subsubsection{Short-term Co-occurrence Modeling}
Our method uses attention mechanisms to model the short-term co-occurrence of events. For an event representation $\bm x_i$, we generate an event neighborhood representation $\bm v_i$ based on the attention mechanism and its neighborhood event representations.

We assume that events occurring in a short period are out-of-order, so the short-term co-occurrence between events can be captured using the attention mechanism that does not consider the order.

Then we introduce the short-term co-occurrence modeling mechanism in detail. For the event $e_i$, we consider that the events occur within the time interval $(e_i.t - T_r, e_i.t + T_r)$ have a short-term co-occurrence with $e_i$. We use $Ne(i)$ to represent the index set of these events. $Ne(i)$ is defined as follows:
\begin{equation}
\setlength{\abovedisplayskip}{3pt}
Ne(i) = \{j|\ \vert e_j.t - e_i.t\vert \leq T_r\}\label{1}
\setlength{\belowdisplayskip}{3pt}
\end{equation}
Referring to the attention mechanism in related works\cite{vaswani2017attention}, out method calculates $v_i$ as follows:
\begin{equation}
\setlength{\abovedisplayskip}{3pt}
\bm v_i = Attention(i, Ne(i)) = \sum_{j \in Ne(i)}a_{ij}(W_v\bm{x}_j)\label{2}
\setlength{\belowdisplayskip}{3pt}
\end{equation}

\begin{equation}
\setlength{\abovedisplayskip}{3pt}
a_{ij} = Softmax_j(s_{ij}) = \frac{exp(\hat{s}_{ij})}{\sum_{j \in Ne(i)}exp(\hat{s}_{ij})}\label{3}
\setlength{\belowdisplayskip}{3pt}
\end{equation}

\begin{equation}
\setlength{\abovedisplayskip}{3pt}
\hat{s}_{ij} = \left \{
    \begin{array}{lr}
    s_{ij},  &\text{if $s_{ij}$ is greater than the topk-th}\\
            &\text{greatest number in $\{s_{ij} | j \in Ne(i)\}$}\\
    -\infty, &\text{otherwise} 
    \end{array}
\right.\label{4}
\setlength{\belowdisplayskip}{3pt}
\end{equation}
where $s_{ij} = q_i^Tk_j$, $q_i = W_q \bm{x}_i$, $k_j = W_k\bm{x}_j$.

$W_v, W_q, W_k \in \mathbb{R}^{D_a \times D_m}$, where $D_a$ is the dimension of the attention mechanism. 

As $\vert Ne(i)\vert$ could be quite large in real data, it is difficult to capture all of the co-occurrences. Thus for an event $e_i$, we only capture $topk$ events which are the closest. 

Our method also use a multi-head attention mechanism. The final representation $\bm v_i$ is as follows:
\begin{equation}
\setlength{\abovedisplayskip}{3pt}
\bm v_i = Concat(head_1, head_2, \cdots, head_{N_h})\label{5}
\setlength{\belowdisplayskip}{3pt}
\end{equation}

Where $head_j = Attention_j(i, Ne(i))$ is defined above. And each $Attention_j$ shares the same structure but has separate parameters $W_v^j, W_q^j, W_k^j$. We guarantee that $D_a \times N_h = D_m$.
\subsubsection{Interval-based Event Pooling}
$\bm v_i$ contains neighborhood information around $e_i$. If the event $e_i$ is very close to $e_j$, the information contained in $\bm v_i$ and $\bm v_j$ will be quite similar. Due to the similarity of the neighboring element information, it is difficult to directly process the sequence $[v_1, v_2, \cdots, v_L]$ with RNN.

So we propose a pooling mechanism based on event interval to solve the above problems. In this paper, the clinical event sequence is first divided into several non-overlapping event groups according to the distribution density of events, and then each group goes through a max-pooling layer separately. The division should satisfy two conditions: 1) the period covered by an event group must be as small as it can; 2) the number of event groups should not be too large.

Let $G_i = \{k| e_k \in Group_i\}$ be the index set of the events contained in the i-th event group, $\{G_i\}$ is the set of all event groups.

$M$ is the group number limit and $L_G$ is the limit number of events in a group.

To make each group's time span as small as it can, we define time span function as follows:
\begin{equation}
\setlength{\abovedisplayskip}{3pt}
span(G_i) = \max_{j, k \in G_i}\{e_j.t - e_k.t\}\label{6}
\setlength{\belowdisplayskip}{3pt}
\end{equation}

The optimal partition of the sequence can be obtained by minimizing the maximum time span of the partition:
\begin{equation}
\setlength{\abovedisplayskip}{3pt}
argmin_{\{G_i\} }\max_{G_i \in \{G_i\}}span(G_i)\label{7}
\setlength{\belowdisplayskip}{3pt}
\end{equation}

We can get the optimal partition with dichotomy and greed algorithm. 

After the max-pooling in each event group, the representation of the event group $\bm g$ can be obtained. The representation $\bm g_i$ of the i-th event group can be calculated as follows:
\begin{equation}
\setlength{\abovedisplayskip}{3pt}
\bm g_i = \max_{i' \in G_i}\bm{v}_{i'}\label{8}
\setlength{\belowdisplayskip}{3pt}
\end{equation}

\subsubsection{Long-term Temporal Dependency Modeling}
We use LSTM to deal with event group representation sequence. In t-th iteration, LSTM cell takes former output $\bm h_{t-1}$, state $\bm c_{t-1}$ and the event group representation sequence input $\bm g_t$ as input and generate $\bm h_t$ as output.
\begin{equation}
\setlength{\abovedisplayskip}{3pt}
\bm h_t = LSTM(\bm{h}_{t - 1}, \bm{c}_{t - 1}, \bm{g}_t)\label{9}
\setlength{\belowdisplayskip}{3pt}
\end{equation}
$LSTM(\cdot)$ represents an iteration. 

The last output $\bm{h}_M$ is the representation for the clinical event sequence as well as the patient.

We use a sigmoid function to get the prediction $\hat{y}$ from the patient representation $h_M$:
\begin{equation}
\setlength{\abovedisplayskip}{3pt}
\hat{y} = \sigma(W_p\bm{h}_M + b_p)\label{10}
\setlength{\belowdisplayskip}{3pt}
\end{equation}

$W_p \in \mathbb{R}_{D_m}$, $b_p \in \mathbb{R}$ are the parameters to learn.

Then we use a cross entropy loss function to calculate the classification loss from the true label $y$ and the prediction $\hat{y}$:
\begin{equation}
\setlength{\abovedisplayskip}{3pt}
Loss(\hat{y}, y) = -\big(y \times ln(\hat{y}) + (1 - y) \times ln(1 - \hat{y})\big)\label{11}
\setlength{\belowdisplayskip}{3pt}
\end{equation}

\section{Experiments}
\subsection{Experiment settings}
This part describes the parameter settings and model training methods of the MRM proposed in this paper.

The parameter settings for MRM are as follows:

Model dimension $D_m$ is 64, refined event number $N_c$ is 3418, feature number $N_f$ is 649 and maximum feature number of a event is 3. The time interval $T_r$ is 0.5 hour, the attention number $N_h$ is 8, the dimension of the attention mechanism is 8 and the number of reserved events $topk$ is 4. The maximum number of the event group $M$
is 64 and the maximum length of the event group is 32.

This work divides the dataset into 3 parts: training set (70\%), validation set (10\%), and test set (20\%). 

All of the network structures mentioned in this part are implemented in Keras and Theano and optimized with the Adam method.

\subsection{Experiment analysis}

We compare MRM proposed in this paper with two types of models: traditional statistical models and sequential neural network models. We use the two datasets, death and labtest described in previous work\cite{liu2019learning}.

The sequential models use the output of the event representation described in Chapter III(B) as its input. The statistical models use a vector $FV$ which records the number of event occurrences as its input. $FV$ is defined by $
FV = \sum^L_{i = 1} \tilde{x}^c_i\label{28}
$, where $\tilde{x}^c_i$ is the one-hot encoding vector for the event $e_i$.

The following is the baseline models of MRM:

\begin{itemize}
\item SVM takes $FV$ vector as its input. 

\item Logistic Regression takes $FV$ vector as its input and adds L2 regularization layer. It is noted as LR. 

\item LSTM uses the LSTM model to process event sequential data and adds a sigmoid layer for prediction at the end. 

\item RETAIN is described in related work. This method uses a fixed partition of length 32 to partition the sequence.  

\item Timeline is described in related work. The input configuration is the same as RETAIN. 

\item TCN is described in related work.
\end{itemize}

\begin{table}
\caption{Performance compared with baselines}
\centering
\begin{tabular}{c c c c c}  
\hline
\textbf{Methods} & \textbf{AUC(death)} & \textbf{AP(death)} & \textbf{AUC(labtest)} & \textbf{AP(labtest)}\\
\hline
SVM & 0.7523 & 0.5154 & 0.6587 & 0.2987 \\
LR & 0.8843 & 0.5213 & 0.6839 & 0.3014 \\
\hline
RETAIN & 0.8967 & 0.6244 & 0.7325 & 0.3196 \\
Timeline & 0.9349 & 0.7119 & 0.7455 & 0.3456 \\
LSTM & 0.9455 & 0.7414 & 0.7495 & 0.3513 \\
TCN & 0.8752 & 0.5752 & 0.7234 & 0.3131 \\
\hline
MRM & \textbf{0.9512} & \textbf{0.7695} & \textbf{0.7688} & \textbf{0.3714} \\
\hline
\end{tabular}

\label{tab:performance}
\end{table}

Table \ref{tab:performance} shows the experimental results of each model on two datasets. Based on the experimental results in table \ref{tab:performance}, we can draw the following conclusions:

\begin{itemize}
\item All sequential models perform better on both tasks than SVM and LR which are based on event frequency. This is because SVM and LR not only ignore the temporal information of the events but also ignore the feature information of the events.

\item RETAIN and TCN perform poorly in both tasks. Although RETAIN and TCN both use a multi-level representation to model clinical event sequences, their division of events depends either on the visit information that exists in the data or on fixed step size.

\item The MRM model proposed in this paper overperforms other models in both tasks. On the death prediction dataset, MRM increased by at least 0.6\% on the AUC indicator and by 3.7\% on the AP indicator relative to other models. On the potassium ion concentration abnormality detection dataset, MRM increased by 2.5\% on the AUC indicator and by 5.7\% on the AP indicator. This is because MRM models the short-term co-occurrence of events with attention mechanism and reduces the length of the sequence by the pooling mechanism, which reduces the difficulty of long-term temporal dependency capture.
\end{itemize}

\section{Conclusion}
We propose a multi-level representation model MRM for long clinical event sequences generated from EHR with complex event types and multi-scale temporal information. MRM uses a sparse attention mechanism to capture the short-term co-occurrence of events and uses interval-based event pooling mechanism to reduce sequence length and to preserve as much the temporal information between events as possible. Experiments on the death prediction dataset and the potassium ion concentration abnormality detection dataset constructed on the open dataset MIMIC-III have proved the effectiveness of the MRM. 

\section{Acknowledgement}
This paper is partially supported by
National Key Research and Development Program of China with Grant No. 2018AAA0101900, Beijing Municipal Commission of Science and Technology under Grant No. Z181100008918005, and the National Natural Science Foundation of China (NSFC Grant No. 61772039 and No. 91646202).

\bibliographystyle{IEEEtran}
\bibliography{citation}

\end{document}